\documentclass{article}

\usepackage{microtype}
\usepackage{graphicx}
\usepackage{subfigure}
\usepackage{booktabs}
\usepackage{hyperref}
\usepackage[accepted]{icml2025}  % none, accepted
\usepackage{amsmath,amsfonts,bm,upgreek,cancel,nicefrac}

\newcommand{\e}{\mathrm{e}}
\newcommand{\E}{\mathbb{E}}

\newcommand{\R}{\mathbb{R}}
\newcommand{\dataset}{\mathcal{D}}

\newcommand{\roi}{\mathcal{R}}

\newcommand{\rx}{\textnormal{x}}
\newcommand{\ry}{\textnormal{y}}

\newcommand{\rva}{\mathbf{a}}

\newcommand{\rvx}{\mathbf{x}}

\newcommand{\va}{\bm{a}}
\newcommand{\vb}{\bm{b}}

\newcommand{\vx}{\bm{x}}

\newcommand{\gF}{\mathcal{F}}
\newcommand{\gG}{\mathcal{G}}

\usepackage{amssymb}
\usepackage{mathtools}
\usepackage{amsthm}
\theoremstyle{plain}

\theoremstyle{definition}

\theoremstyle{remark}

\usepackage{doi}            % doi hyperlinks

\icmltitlerunning{Integrated Gradient Correlation}

\begin{document}

\twocolumn[
\icmltitle{Integrated Gradient Correlation:\\ a Dataset-wise Attribution Method}

\icmlsetsymbol{git}{*}

\begin{icmlauthorlist}
\icmlauthor{Pierre Lelièvre}{git,ntu}
\icmlauthor{Chien-Chung Chen}{ntu}
\end{icmlauthorlist}

\icmlaffiliation{ntu}{Department of Psychology, National Taiwan University, Taipei, Taiwan}

\icmlcorrespondingauthor{Pierre Lelièvre}{contact@plelievre.com}

\icmlkeywords{Machine Learning, ICML, Deep Learning, Attribution, Interpretation, Explanation, Neuroscience, Neural Representation, fMRI, pRF}

\vskip 0.3in
]

\printAffiliationsAndNotice{$^*$\href{https://github.com/plelievre/int_grad_corr}{Git repository}}

\begin{abstract}
    Attribution methods are primarily designed to study input component contributions to individual model predictions. However, some research applications require a summary of attribution patterns across the entire dataset to facilitate the interpretability of the scrutinized models at a task-level rather than an instance-level. It specifically applies when the localization of important input information is supposed to be \textit{stable} for a specific problem but remains unidentified among numerous components. In this paper, we present a dataset-wise attribution method called Integrated Gradient Correlation (IGC) that enables region-specific analysis by a direct summation over associated components, and further relates the sum of all attributions to a model prediction score (correlation). We demonstrate IGC on synthetic data and fMRI neural signals (NSD dataset) with the study of the representation of image features in the brain and the estimation of the visual receptive field of neural populations. The resulting IGC attributions reveal selective patterns, coherent with respective model objectives.
\end{abstract}

%-------------------------------- Main Matter

\section{Motivation}\label{sec:motivation}

Existing attribution methods study input component contributions to individual model predictions. Here, we investigate the problem of summarizing these attributions at a dataset level. We particularly search for a method fulfilling the interpretability requirements of modeling scenarios such as \ \textit{Where a specific image feature is represented in the visual cortex of the human brain?} \textit{What is the visual receptive field of a population of neurons?} (see Section.\ref{sec:applications} for details). These examples have in common to call for the exposition of which input components are responsible for achieving the overall task, and not specific entries. An implicit condition of these scenarios is thus a relatively \textit{stable} localization of important input information across the dataset, at least for each specific task. For instance, it works with functional Magnetic Resonance Imaging (fMRI), where each voxel/vertex represents an invariant brain area, covering a population of neurons. On the other hand, in the context of image content recognition, a \textit{dataset-wise attribution method} would not be pertinent because a \textit{cat} can appear anywhere in an image and still accurately trigger the \textit{cat} class. While such a requirement can be perceived as a limitation, this is an intrinsic aspect of any dataset-wise inquiry (that do exist), and not of our resolving method.

Some research fields, such as neuroscience, heavily rely on linear models to straightforwardly associate prediction scores with input components. Modelers overcome inherent limitations with \textit{linearized} data, based on handcrafted nonlinear data preprocessing. Although legitimate, such an approach is unlikely to produce perfect linear mappings. Only deep models could exploit nonlinear and multiscale interactions without \textit{a priori} data domain adaptation. However, the lack of dataset-wise summary techniques leaves deep models with an interpretability issue.

Our method is therefore designed to be easily integrated in research activities and transparently used in place of linear regression analysis. To do so, it must fulfill requirements expressed in \citet{NaselarisEncodingdecodingfMRI2011} with the following questions: \textit{Does an input region of interest (ROI), contain information about some specific set of output features? Are there specific ROIs that contain relatively more information about a specific set of features? Are there specific features that are preferentially represented by a single ROI?} We then identify three main specifications: \textbf{1)} a \textit{dataset-wise attribution method} must permit a flexible definition of ROIs; \textbf{2)} relative ROI attributions must allow pertinent comparisons; \textbf{3)} dataset-wise attributions must enable comparisons between different features, and implicitly for identical features, but through different models.

\subsection{General modeling context}

We define deep models as a class of functions $f:\R^m\to\R$ by $\gF$. These models predict a random variable $\ry \in \R$ from a random vector $\rvx\in\R^m$, and they are trained with $n$ i.i.d pairs of a dataset $\dataset$, so that $(\rvx,\ry)\sim\dataset = \{(\vx^{(i)},y^{(i)})\}^n_{i=1}$.

%----------------

\section{Attribution methods for individual predictions}

Available techniques for individual predictions arose from different research domains. For instance, the Shapley Value \citep{ShapleyValueNPersonGames1952} originated from the cooperative game theory. It addressed cost/gain sharing problems between players w.r.t\ their contributions to the outcome. Modern variants \citep{LundbergUnifiedApproachInterpreting2017, DhamdhereShapleyTaylorInteraction2020, SundararajanmanyShapleyvalues2020} reduced inherent computational cost, but it is still not suited for large input dimensionalities. More recently, image content recognition deep models prompted the development of methods displaying pixel arrangements responsible for a classification decision \citep{ZeilerVisualizingUnderstandingConvolutional2013}. Being differentiable, these models led to techniques exploiting gradient back-propagation for a lower computational cost \citep{BaehrensHowExplainIndividual2009,SimonyanDeepConvolutionalNetworks2014}. However, naive gradients encounter issues with activation functions such as ReLU, as it back-propagates zero gradients for negative inputs. Consequently, methods like guided back-propagation \citep{SpringenbergStrivingSimplicityAll2015}, GradCAM \citep{SelvarajuGradCAMVisualExplanations2016,ChattopadhyayGradCAMImprovedVisual2018}, LRP \citep{BinderLayerwiseRelevancePropagation2016}, or DeepLift \citep{ShrikumarLearningImportantFeatures2019} operated architectural tweaks to avoid attribution shortcomings. Other methods also include \citet{ZhouLearningDeepFeatures2015,RibeiroWhyShouldTrust2016,ZintgrafVisualizingDeepNeural2017}.

An alternative to prevent spurious gradients is to compare attributions of the input under scrutiny with a baseline. This way, even if some input components receive zero gradients, their comparison with possibly non-zero baseline gradients reveals more correct contributions. Among popular \textit{baseline attribution methods}, we find \textit{path methods} and particularly \textit{Integrated Gradients} (IG) \citep{SundararajanAxiomaticAttributionDeep2017}. Based on \citet{AumannValuesnonatomicgames1974}, IG aggregates the gradients of linearly interpolated inputs, following the shortest line from the baseline to the input (with all component attributions updated at once for few forward/backward passes through the model). For instance, another \textit{path method} like Baseline Shapley (BS) \citep{SundararajanmanyShapleyvalues2020} requires successive activations of each input component, making it longer to compute than IG by several folds. Although popular, \textit{path methods} still exhibit unresolved issues, such as the optimality of selected path \citep{KapishnikovGuidedIntegratedGradients2021,SanyalDiscretizedIntegratedGradients2021,KawaiCompensatedIntegratedGradients2022} or the choice of the most appropriate baseline \citep{SmilkovSmoothGradremovingnoise2017,FongInterpretableExplanationsBlack2017,LundbergUnifiedApproachInterpreting2017,ErionImprovingperformancedeep2020,SturmfelsVisualizingImpactFeature2020,XuAttributionScaleSpace2020,PanExplainingDeepNeural2021,TanMaximumEntropyBaseline2022,LiuRethinkBaselineIntegrated2023}.

Our definition of \textit{dataset-wise attribution methods} is built upon existing methods for individual predictions. It particularly requires desirable properties (axioms) that are only satisfied by \textit{path methods} (see Section.\ref{sec:dw_attr_meth}). We will therefore briefly detail this class of methods, while referring to \citet{LundstromRigorousStudyIntegrated2022} for an in-depth analysis.

\subsection{Path methods}

A function $\gamma:\R^m\times\R^m\times[0,1]\to\R^m$ is a \textit{path function} if for any fixed input $\vx^{(i)}$ and baseline $\bar{\vx}$, $\gamma$ is a continuous and piece-wise smooth curve from $\bar{\vx}$ to $\vx^{(i)}$ over $t$. This class of functions is represented by $\gG$, and intermediary inputs are defined as $\vx^{(t)} = \gamma(\vx^{(i)}, \bar{\vx}, t)$.

Given functions $f$ and $\gamma$, fixed input $\vx^{(i)}$ and baseline $\bar{\vx}$, a \textit{path method} is a function of the form $a:\gF\times\gG\times\R^m\times\R^m\to\R^m$. The contribution of $\vx^{(i)}$ components to the individual prediction $f(\vx^{(i)})$ are denominated by the vector $\va^{(i)} = a(f,\gamma,\vx^{(i)},\bar{\vx})$, which $j$-th component is given by:
\begin{equation}\label{eq:path_method}
    \va_j^{(i)} = \int_0^1 \frac{\partial f}{\partial \vx_j^{(t)}}(\gamma(\vx^{(i)}, \bar{\vx}, t)) \times \frac{\partial \gamma_j}{\partial t}(\vx^{(i)}, \bar{\vx}, t) dt
\end{equation}
Function $f$ is already supposed to be differentiable because deep models are trained \textit{via} SGD-like algorithms. However, \textit{path methods} further imply that partial derivatives $\nicefrac{\partial f}{\partial \vx_j^{(t)}}$ exist for all $\vx^{(t)}$ \textit{almost everywhere}.\footnote{
    It means that the points for which partial derivatives are not defined have Lebesgue measure 0.
}

The baseline $\bar{\vx}$ is similar to $\vx^{(i)}$ in terms of dimensionality, but it can be arbitrarily set to any value. Zeros can simulate the least informational input in many cases, e.g., a black image. Nonetheless, defining a relevant baseline is difficult in some scenarios. For instance, the activity of the brain in a resting state is rarely flat to zero. In such situations, we can turn the baseline into a random vector $\bar{\rvx}$ sampled from any distribution, and further consolidate attributions by taking their expected value \citep{LundbergUnifiedApproachInterpreting2017,ErionImprovingperformancedeep2020}. The choice of the baseline distribution is commonly the dataset $\dataset$ itself, so that:
\begin{equation}
    a_j(f,\gamma,\vx^{(i)},\bar{\rvx}) = \E_{\bar{\rvx}\sim\dataset} \left[a_j(f,\gamma,\vx^{(i)},\bar{\vx}^{(k)})\right]
\end{equation}
\textit{Random baseline path methods} fulfill a similar set of axioms as original \textit{path methods}, and the following definitions could be easily adapted to this variant. We use it in practice, but we will keep the single baseline paradigm for legibility.

\paragraph{Dummy} This first axiom intuitively means that an input component having no influence on model outputs should have zero attribution. In other word, if a partial derivative of $f$ is zero for a component $j$, corresponding $\va_j^{(i)}$ is null.  % Also called \textit{sensitivity(b)}

\paragraph{Completeness} In a cost/gain sharing context, \textit{completeness} refers to the intuitive idea that the sum of all contributions must be equal to the cost/gain under scrutiny. More generally, it means that the sum of all component attributions $\va_j^{(i)}$ must reflect the model prediction $f(\vx^{(i)})$, so that:
\begin{equation}\label{eq:completeness}
    \sum_{j=1}^m \va_j^{(i)} = f(\vx^{(i)}) - f(\bar{\vx})
\end{equation}

\paragraph{Implementation invariance} Two functionally equivalent models (i.e.\ exhibiting comparable input/output behaviors), but with different architectures (e.g., the number of layers or the type of activation functions), should produce similar attributions. For instance, attribution methods that require low-level model modifications depending on selected activation functions, such as DeepLIFT \citep{ShrikumarLearningImportantFeatures2019}, break this axiom. In addition, the status of implied architectural changes becomes unclear when attempting to explain the default behavior of these models.\footnote{
    \textit{Path methods} also satisfy \textit{linearity} (see Appendix.\ref{sec:linearity}), but this axiom is not necessary for the definition of our method.
}

%----------------

\section{Dataset-wise attribution methods}\label{sec:dw_attr_meth}

We define \textit{dataset-wise attribution methods} as functions summarizing the distribution of component attributions for individual predictions $\va^{(i)}_j$ collected over the whole dataset $\dataset$ and represented by the random vector $\rva_j$. However, the isolated dynamic of $\rva_j$, e.g., its mean or SD, is usually irrelevant (see Section.\ref{sec:benchmark} for illustrations). From the \textit{completeness} axiom of \textit{path methods}, we know that the overall magnitude of individual attributions is linearly modulated by the magnitude of model predictions $f(\rvx)$. We therefore propose to summarize the distribution of $\rva_j$ by a measure of its linear relationship with predictions $f(\rvx)$, i.e.\ a covariance-based analysis.

In addition, the third requirement described in Section.\ref{sec:motivation} states that dataset-wise attributions must be comparable between models and thus implicitly reflect models' performance. The covariance between $\rva_j$ and \textit{true} output $\ry$ then appears an even more informative alternative. In the case of unavailable \textit{true} output $\ry$, we still explore the first option mentioned above in Subsection.\ref{sec:igac}. Finally, we introduce a scaling factor in our definition to enable interesting properties (see below).\footnote{
    Despite the absence of apparent alternatives fulfilling the following properties, we leave the uniqueness of our proposition at a state of conjecture.
}

Given functions $f$ and $\gamma$, random variables $\rvx$ and $\ry$, and baseline $\bar{\vx}$, a \textit{dataset-wise attribution method} is a function of the form $b:\gF\times\gG\times\R^m\times\R^m\times\R\to\R^m$. Resulting attributions are denominated by the vector $\vb = b(f,\gamma,\rvx,\bar{\vx},\ry)$, which $j$-th component is given by:
\begin{equation}\label{eq:igc}
    \vb_j = \frac{1}{\sigma_{f(\rvx)}\sigma_\ry} \E_{(\rvx,\ry)\sim\dataset}\left[\va_j^{(i)}\times(y^{(i)} - \mu_{\ry})\right]
\end{equation}
with $\sigma_{f(\rvx)}$ and $\sigma_\ry$ the standard deviations of $f(\rvx)$ and $\ry$, and $\rva_j$ attributions given by a \textit{path method} (see Eq.\ref{eq:path_method}).\footnote{
    We omit the subtraction of $\va_j^{(i)}$ by the expected value of $\rva_j$ for legibility and the simplification of the \textit{completeness} proof.
}

\paragraph{Dummy} Similarly to \textit{path methods}, our proposition is zero for an input component if the partial derivatives of $f$ is zero for this component. In addition, \textit{dummy} can be extended to input component for which attributions $\rva_j$ are independent of $\ry$, i.e.\ having zero covariance.

\paragraph{Implementation invariance} This property is inherited from the supporting \textit{path method}.\footnote{
    We empirically demonstrate the relative independence of our method upon model implementations in Appendix.\ref{sec:benchmark_supp}.
}

\paragraph{Completeness to a model prediction score}

The first two interpretability requirements expressed in Section.\ref{sec:motivation} impose that dataset-wise attributions can be easily aggregated. For any ROI $\roi$, the summary $\vb_\roi$ is thus defined as a simple summation over input components.
\begin{equation}\label{eq:additive_property}
    \vb_\roi = \sum_{j\in\roi} \vb_j
\end{equation}
Similarly to the \textit{completeness} axiom of \textit{path methods}, such an additive property is only meaningful if the total attribution over input components reflects an interesting characteristic of the model. Thanks to scaling $\nicefrac{1}{\sigma_{f(\rvx)} \sigma_\ry}$, our proposition sums to the correlation $\rho\left(f(\rvx),\ry\right)$ (see Appendix.\ref{sec:igc_completeness_proof} for the proof). The total attribution is therefore related to a prediction score of the model, and fulfills the third requirement expressed in Section.\ref{sec:motivation}.
\begin{equation}
    \sum_{j=1}^m \vb_j = \rho\left(f(\rvx),\ry\right)
\end{equation}

\subsection{Supporting \textit{path method}}

Each \textit{path method} provides different attributions since they take different paths from the baseline to the input. However, empirical results with two popular \textit{path methods} (BS and IG) showed that BS does not produce significant differences from IG (see Appendix.\ref{sec:benchmark_supp}). Because of BS higher computational cost, we recommend IG in the general case. This is also the reason we coined our \textit{dataset-wise attribution method}: Integrated Gradient Correlation (IGC).

By definition, IG is a \textit{path method} with $\gamma^{\text{IG}}(\vx^{(i)}, \bar{\vx}, t)=\bar{\vx} + t (\vx^{(i)}-\bar{\vx})$, i.e.\ a straight line between $\bar{\vx}$ and $\vx^{(i)}$. In practice, the integral of Eq.\ref{eq:path_method} is replaced by its Riemman approximation with few discrete $s$ steps \citep{SundararajanAxiomaticAttributionDeep2017}:
\begin{equation}\label{eq:int_grad}
    \va_j^{(i)\ \text{IG}} \approx \frac{\vx_j^{(i)} - \bar{\vx}_j}{s} \sum_{t=1}^s \frac{\partial f}{\partial \vx_j^{(t)}} \big(\bar{\vx} + \frac{t}{s} (\vx^{(i)} - \bar{\vx})\big)
\end{equation}

\subsection{Auto-correlation variant}\label{sec:igac}

IGC is primarily designed to explain a model and its associated dataset, but \textit{true} outputs $\ry$ may not be available when applied on new data. In this case, we propose an \textit{auto-correlation} variant $\vb^\ast$ (IGaC), where $\ry$ is replaced by $f(\rvx)$ in Eq.\ref{eq:igc}, so that:
\begin{equation}\label{eq:igac}
    \vb_j^\ast = \frac{1}{\sigma_{f(\rvx)}^2} \E_{\rvx\sim\dataset}\left[\va_j^{(i)}\times\big(f(\vx^{(i)}) - \mu_{f(\rvx)}\big)\right]
\end{equation}
IGC properties are conserved except for \textit{completeness} that sums to 1 instead of a model prediction score.
\begin{equation}
    \sum_{j=1}^m \vb_j^\ast = \rho\left(f(\rvx),f(\rvx)\right) = 1
\end{equation}

\subsection{IGC with categorical distributions}\label{sec:cat_dist}

Our method is mainly elaborated for models predicting scalars. However, some modeling scenarios involve the prediction of random variables ruled by categorical distributions with $K$ categories. Usually expressed as \textit{one-hot} vectors, models predict probabilities $p_k$ with $k\in[1,K]$. A prediction score based on the correlation between $p_k$ and corresponding \textit{true} \textit{one-hot} vector components is not ideal, but effective (see Exp.D in Section.\ref{sec:benchmark}). In addition, this setup allows the computation of dataset-wise attributions w.r.t.\ each class independently.

\subsection{Practical IGC computation}\label{sec:practical_igc}

To be dataset-wise relevant, IGC attributions require a representative (large enough) validation set. Conversely, the computation time of our method linearly scales up with the size of this set, and smaller sub-sets may be necessary in some contexts (e.g., extremely large datasets, limited computational resources). Under those circumstances, we advise checking if the absolute difference between the total sum of IGC attributions (computed on a subset) and the model correlation score (computed on the complete validation set) remains below an acceptable value. We used $1\e{-2}$ in our experiments.

Nonetheless, reaching this figure necessitates relevant IG entries, computed with a number of steps (and random baselines) appropriate to the model complexity. Based on \textit{completeness} (see Eq.\ref{eq:completeness}), we advise checking whether the mean absolute difference between the sum of IG components and its corresponding input/baseline difference is reasonable ($<1\e{-3}$ in our studies).

%----------------

\section{Benchmark}\label{sec:benchmark}

To our knowledge, there is no existing \textit{dataset-wise attribution method} for comparison with IGC. The only available baselines are naive aggregations of IG attributions, i.e.\ the mean and SD of their per-component distributions over the dataset.\footnote{
    In Appendix.\ref{sec:benchmark_supp}, we also report basic metrics not involving any model, such as per-component correlations and t-tests.
} In addition, available quantitative methods to evaluate attribution methods for individual prediction like LeRF\&MoRF \cite{SamekEvaluatingvisualizationwhat2015} or ROAR \cite{HookerBenchmarkInterpretabilityMethods2019} are designed for classification endeavors. They require an ordering of input components from least to most relevant. It works for categorical tasks because an input component with a negative attribution indicates that it lowers the output probability of this class and thus increases the model uncertainty. On the contrary, for models predicting scalars, a negative attribution is not necessarily the least relevant, e.g., when the output is also negative.

\begin{figure*}[t]
    \centering
    \includegraphics[width=\linewidth]{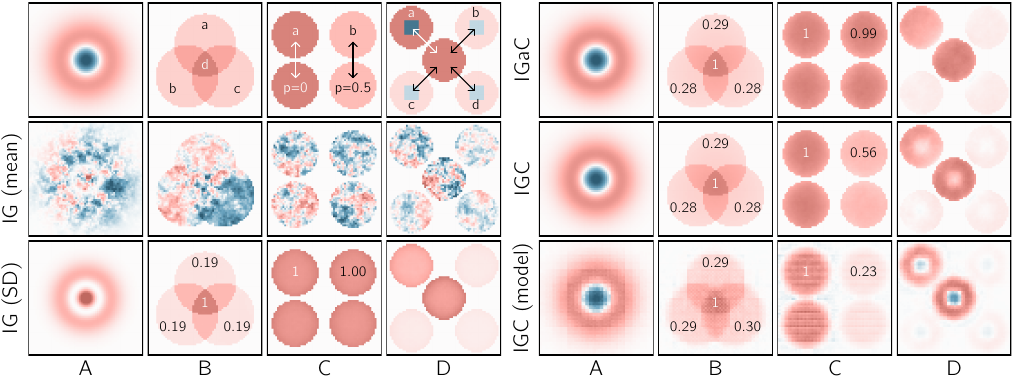}
    \caption{Benchmark of different propositions of \textit{dataset-wise attribution methods} on four experiments labeled A-D. Attribution maps are individually scaled for a better visualization. Blue/red pixels indicate negative/positive values. See the main text for details about the experiments and for a discussion of displayed propositions.}
    \label{fig:benchmark_main}
\end{figure*}

In the current work, we therefore assess the relevancy of IGC attributions on synthetic experiments designed to reveal shortcomings of other propositions. We first define localized image statistics, e.g., the sum of pixel values weighted by a 2-d map, and then try to recover the generating masks/rules from the pairs of input image/output statistic. Images are randomly generated, but we enforce a spatial frequency distribution similar to natural images to guarantee a certain level of spatial redundancy (see Fig.\ref{fig:benchmark_images} image examples and Appendix.\ref{sec:benchmark_images} for implementation details). To prevent implementation and training variabilities of deep models, presented attributions are computed through the \textit{original} differentiable functions producing the statistics. Only \textit{IGC (model)} attributions involve trained models for verification.\footnote{
    R2 scores of the models dedicated to Exp.A,B,C are $>0.99$, $>0.99$, and $0.81$. For Exp.D, the Top-1 accuracy is 0.81. Statistics for Exp.C,D involve some randomness naturally limiting maximum prediction scores. Appendix.\ref{sec:benchmark_model_details} reports architectural/training characteristics of implemented models and IGC computation details.
} The definition of the four experiments labeled Exp.A-D are detailed below and illustrated in Fig.\ref{fig:benchmark_main} (top/left row).

\vspace{-2mm}  % Improve general layout.
\paragraph{Experiment.A}

This is the sum of pixel values weighted by the difference of two centered Gaussians with opposite polarities. From Fig.\ref{fig:benchmark_main}, it is clear that the mean of IG attributions is not suitable. It only produces residual noise.\footnote{
    Regions outside masks are perfectly zero because we use of gradients from original functions to compute the different attribution propositions. However, this is of little concern as observed shortcomings happen inside the masks.
} Indeed, as output values are standardized (as it is usually done to train deep models), \textit{IG (mean)} attributions tend to be zero by \textit{completeness}. For instance, if we consider a trivial and perfect model $f$ predicting a standardized random scalar $\rx$ from the same input $\rx$ (i.e.\ learning identity), the IG value for any $x^{(i)}$ would be $x^{(i)}$, and \textit{IG (mean)} attributions would be $\E_{\rx\sim\dataset}[x^{(i)}] = 0$. Outcomes from \textit{IG (SD)} look more appropriate. However, this alternative makes is incapable of negative attributions by definition. \textit{IG (SD)} is thus missing important information that could lead to model strategy misinterpretations. Only IGC, IGaC, and \textit{IGC (model)} capture the expected attribution pattern.

\paragraph{Experiment.B}

We first compute the mean value of three independent circular regions, and then report the maximum value. The expected dataset-wise attribution is then the probability of each pixel to contribute to the output. As circles are partially overlapped, probabilities must sum at intersections. Once again, only IGC, IGaC, and \textit{IGC (model)} capture the right balance of each area. For instance, the sum of individual densities (a, b, c) represent 85\% of the central area (d) for IGC, where it is only 57\% for \textit{IG (SD)}.

\paragraph{Experiment.C}

The goal of Exp.C is to demonstrate the benefice of IGC over IGaC when there are large discrepancies between models predictions and \textit{true} outputs. To simulate this phenomenon, we introduced some randomness during the computation of the statistic, so that when we calculate the attributions displayed in Fig.\ref{fig:benchmark_main}, this initial random process is inaccessible. The statistic itself is computed in two steps. We first record the cosine similarity between two vertical pairs  of circular regions (a, b), and then report the maximum value. The randomness is introduced in (b), having a probability $p=0.5$ to be set to zero. Taking (a) as a reference, the mean attribution of (b) is similar to (a) for IGaC and about 0.56 for IGC. So, only IGC manages to reflect the uncertainty associated with (b). For \textit{IGC (model)}, the model seems to further adapt to the reliability of the inputs, and (b) even drops to 0.23.

\paragraph{Experiment.D}

With Exp.D, we finally explore categorical outputs. We report the argmax of the cosine similarity between a central circular area and four corner circles. Only, the first category (a) is reported (other classes being symmetrical). In addition, small squares in each corner circle are randomly substituted. Similarly to Exp.C, only IGC reveals unreliable areas (the corner square and the associated pixels in the central circle), for which IGaC is completely blind. With \textit{IGC (model)}, these \textit{spurious} squares are even assigned negative values. Indeed, a component that increases the model uncertainty about a category by lowering corresponding probability deserves a negative attribution.

Overall, IGC proved to be a non-trivial method, able to reveal important patterns that other propositions fail to exhibit. In addition, IGC fulfills \textit{completeness to a model prediction score} that was not covered by this benchmark.

%----------------

\section{Applications}\label{sec:applications}

Integrated Gradient Correlation is applicable to a diversity of real modeling scenarios. Here, we present a decoding model of fMRI data that investigates the representation of image statistics in the brain, and an encoding model that estimates the visual receptive field of neural populations.\footnote{
    Appendix.\ref{sec:application_model_details} reports architectural/training characteristics of implemented models and IGC computation details.
}

\begin{figure*}[t]
    \centering
    \includegraphics[width=\linewidth]{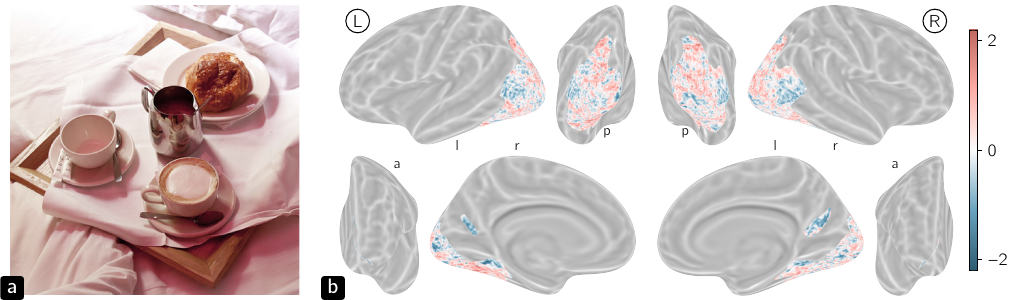}
    \caption{NSD dataset overview. Panel \textbf{a} displays an image as seen by the subject 1, and panel \textbf{b} shows corresponding fMRI activation maps for the left (L) and right (R) brain hemispheres. These surface-based fMRI data are projections from volumetric acquisitions to an average brain space and limited to the visual cortex. Per vertex activations over the dataset are standardized per subjects. For each hemisphere, letters a, l, r, and p denote anterior, left, right, and posterior views.}
    \label{fig:nsd_dataset}
\end{figure*}

\subsection{Representation of image statistics in the brain}\label{sec:fmri_dec}

In neuroscience, fMRI is a widely used technique to measure the brain activity. Neural activations are captured through the proxy of the blood oxygenation level. This intermediary is relatively delayed from the underlying electric neural activity but provides high-resolution maps.

The recent publicly available \href{http://naturalscenesdataset.org}{\textit{Natural Scenes Dataset} (NSD)} \citep{Allenmassive7TfMRI2022} has been designed to enable machine learning inquiries on human vision. It offers fMRI data for $>$70k distinct images, acquired during a long-term recognition task with 8 subjects over one year. Selected images come from the COCO dataset \citep{LinMicrosoftCOCOCommon2015} (see Fig.\ref{fig:nsd_dataset}a). The fMRI data are initially volume-based, with brain hemispheres discretized as voxels, but they can also be projected on the surface of the brain gray matter, which is a convoluted layer where most activations happen. As result, such a surface is easier to map to an average brain morphology shared by all participants (i.e.\ the \textit{fsaverage} template of \href{https://surfer.nmr.mgh.harvard.edu}{\textit{FreeSurfer}} software), and enables inter-subject data aggregation. Here, we focus on the visual cortex, i.e.\ two graphs of nearly 20k vertices each (see posterior areas of left and right hemispheres in Fig.\ref{fig:nsd_dataset}b).

For this first example, we build a model predicting two image statistics from fMRI data. We chose the luminance contrast (i.e.\ the SD of pixel luminance values) which is perhaps the most studied image variable in vision science. The second is \textit{1/f slope}, a value related to spatial frequencies and relevant to our visual perception \citep{FieldRelationsstatisticsnatural1987,TolhurstAmplitudespectranatural1992,TorralbaStatisticsnaturalimage2003}. The slope refers to the decreasing linear correspondence between log-intensities and spatial log-frequencies of natural images. Both features are computed globally on gray-scaled images. Our model is a simple multilayer perceptron (6 layers) with batch-normalization \citep{IoffeBatchNormalizationAccelerating2015} and Mish activation functions \citep{MisraMishSelfRegularized2020}. Despite the morphological mapping to an average brain, the functional behavior of each vertex may differ from one subject to another. As a result, the first layer is optimized per subject and serves as an adaptation interface. The average R2 score is 0.39.

\begin{figure*}[t]
    \centering
    \includegraphics[width=\linewidth]{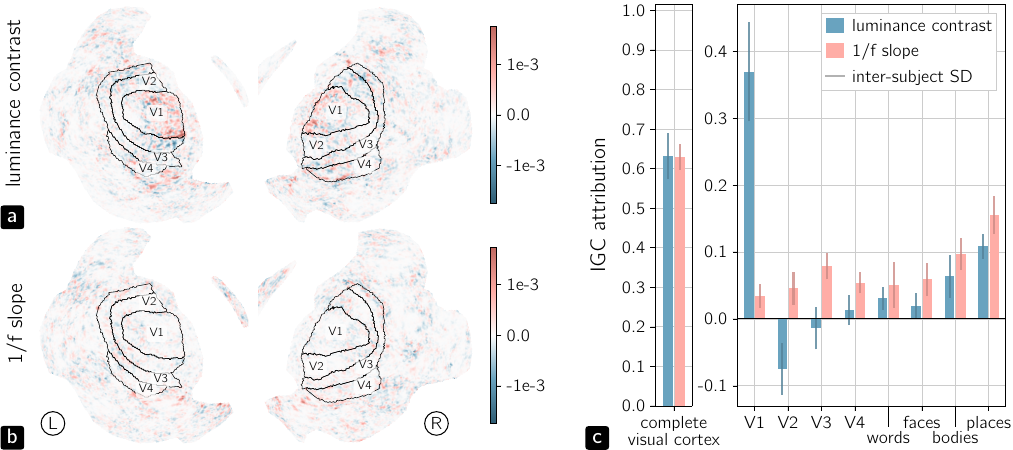}
    \caption{IGC attributions associated with the prediction of image statistics from fMRI data: luminance contrast (panel \textbf{a}) and \textit{1/f slope} (panel \textbf{b}). IGC maps and outlines of early visual ROIs (V1 to V4) correspond to the subject 1 of the NSD dataset. Higher level ROIs (e.g., \textit{bodies} or \textit{faces}) are not represented for legibility because of their scattered nature. Panel \textbf{c} displays an ROI-based summary of IGC attributions. Error bars reflect inter-subject variability (1 SD).}
    \label{fig:igc_imst}
\end{figure*}

Fig.\ref{fig:igc_imst}a,b show resulting IGC attributions for the first subject. In overlay, we display V1 to V4 ROIs that constitute the beginning of the visual pathway in the brain. Early convolutional blocks of image content recognition deep models, such as AlexNet \citep{KrizhevskyImageNetClassificationDeep2012} or VGG \citep{SimonyanVeryDeepConvolutional2014} share some arguable degrees of similarity with these early visual ROIs \citep{YaminsUsingGoalDrivenDeep2016,EickenbergSeeingitall2017,NeriDeepnetworksmay2022}. For the luminance contrast in Fig.\ref{fig:igc_imst}a, we observe that V1 contributions are strongly positive (in red), while V2 and V3 attributions appear negative (in blue). On the other hand, Fig.\ref{fig:igc_imst}b displays a more diffuse activation pattern for \textit{1/f slope}, especially occurring beyond V1 to V4.

Due to the additive property of IGC attributions (see Eq.\ref{eq:additive_property}), a more quantitative analysis is possible by a direct summation over ROIs. Fig.\ref{fig:igc_imst}c then consolidates previously observed trends for all subjects. Nonetheless, how to interpret negative attributions found in V2 for the luminance contrast? Contrary to categorical models, where inputs that systematically negatively contribute to the predictions can be interpreted as increasing the uncertainty of the model, negative IGC values found in models predicting scalars should rather be understood as an adjusting mechanism, balancing an overestimation resulting from positive regions. Therefore, V2 probably serves as a counterbalance to V1, which provides most of the global image luminance contrast estimation. The literature indicates that neurons with large receptive fields and associated with the peripheral vision are already present in V1, so that the strategy revealed by our method seems plausible to reflect real neural mechanisms. Another aspect making us confident about IGC findings, is that higher-level brain areas dedicated to large objects like \textit{bodies} and \textit{places} appear more relevant to the task than ROIs assigned to smaller content such as \textit{words} and \textit{faces}.

Concerning \textit{1/f slope}, IGC demonstrates a contrasting attribution distribution over the visual pathway. Fig.\ref{fig:igc_imst}c presents a more even use of all areas, with a preference for higher-level ROIs. This distinct strategy is in fact pertinent for a statistic summarizing spatial frequencies because the hierarchical processing of multiscale information is an inherent characteristic of our complete visual system.

\subsection{Visual receptive field of neural populations}\label{sec:fmri_enc}

Neurons in early visual areas have been proved to be spatially specialized \citep{EngelRetinotopicorganizationhuman1997,TylerExtendedConceptsOccipital2005,WandellVisualfieldmap2005,WandellVisualFieldMaps2007,MackeyVisualfieldmap2017,BensonHumanConnectomeProject2018}. The population receptive field (pRF) then refers to the localized image area for which a group of neurons is responsive. Hierarchically folded repetitions of the visual field in the brain are even what enable the definition of V1 to V4 ROIs. With a long history in neuroscience, this functional identification of brain regions has been addressed with protocols involving artificial stimuli and a brute-force estimation of model parameters. Instead, we aim at providing pRF data with a wider and more ecological validity than the traditional approach by using our method on a deep encoding model predicting brain signals from natural images.

\begin{figure*}[t]
    \centering
    \includegraphics[width=\linewidth]{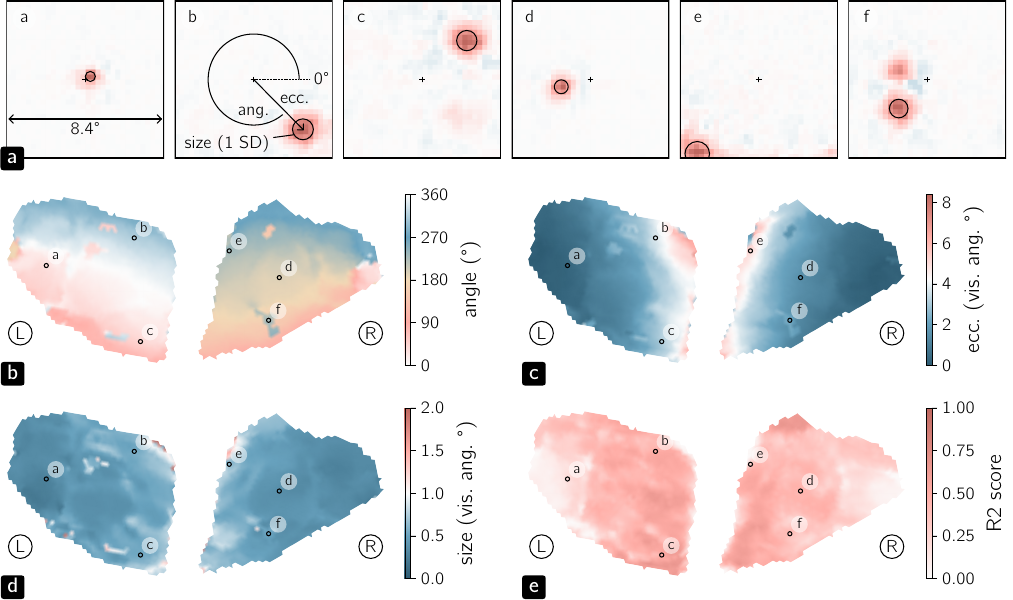}
    \caption{pRF estimation from IGC attributions associated with the prediction of the brain activity (fMRI data) from corresponding image stimuli. Panel \textbf{a} shows a selection of vertices located in V1 of the subject 1 of the NSD dataset. Circles indicate fitted pRF (1 SD) and central small crosses the fixation point of the participants. Panels \textbf{b},\textbf{c},\textbf{d} display pRF parameters (angle, eccentricity, and size) for all vertices of V1. The encoding accuracy R2 of our model is presented in panel \textbf{e}. Vertex labels a-f correspond between panels.}
    \label{fig:igc_prf}
\end{figure*}

Specifically, we use the fMRI data of the NSD dataset restricted to the union of all subjects' V1, and associated images downscaled to 64 pixels. Our model consists of a ConvNeXt convolutional architecture (1 stem + 4 blocks) \citep{LiuConvNet2020s2022}, followed by two fully connected layers using batch-normalization and Mish activation functions. Because of subjects' residual morphological differences (see previous experiment for details), the last linear layer is optimized per subject and serves as an adaptation interface. The average R2 score is 0.36 (see Fig.\ref{fig:igc_prf}e).

For a selection of 6 vertices located in V1 of the subject 1 of the NSD dataset, Fig.\ref{fig:igc_prf}a shows that resulting IGC attributions give a direct visualization of the pRF associated with each vertex. pRF are traditionally modeled by 2-dimensional Gaussian distributions with a center (mean) expressed in polar coordinates, i.e.\ angle and eccentricity (see information overlaid on vertex b). From IGC maps, the fitting procedure is straightforward, and estimated pRF are represented by circles of radius 1 SD. Fig.\ref{fig:igc_prf}b,c,d display fitted parameters of all V1 vertices. Eccentricity and size are expressed in degrees of visual angle. The image size is 8.4°. Overall results are smooth and correspond to our expectations from the literature. Angles depicted in Fig.\ref{fig:igc_prf}b reflect the spatial inversion of images on our retina, e.g., the right side of images is represented in the left hemisphere. In addition, eccentricity and size (Fig.\ref{fig:igc_prf}c,d) exhibit some correlation. pRF are smaller at the center of images corresponding to our fovea. Nonetheless, there are some \textit{spurious} regions, like around vertex f (see Fig.\ref{fig:igc_prf}b). From the IGC map (Fig.\ref{fig:igc_prf}a), we understand that this is due to a bi-modal receptive field that a 2-dimensional Gaussian cannot handle by definition. These anomalies are therefore due to the data preprocessing rather than an inaccuracy of our method.

%----------------

\section{Conclusion}

In this paper, we introduced the concept of \textit{dataset-wise attribution methods} as functions summarizing the distribution of component attributions obtained with existing \textit{path methods} for individual predictions. Our definition is axiomatic with the following desirable characteristics: 1) ROI attributions are computed as the sum of associated components; 2) the total attribution matches a prediction score of the model under inspection. In particular, we use \textit{correlation} as a versatile prediction score and Integrated Gradients as a supporting \textit{path method}. Integrated Gradient Correlation therefore inherits from IG an easy implementation and an applicability to high-dimensional datasets or highly non-linear models, as long as studied models are differentiable and that the number of IG steps is appropriate for the model complexity.

Furthermore, IGC attributions are designed to be \textit{task}-relevant rather than \textit{instance}-relevant. One typical use is when multiple features can be predicted from the same pool of data, and that the respective localization of important input information is still unidentified among numerous components. Such a modeling scheme then naturally meets the only requirement to obtain meaningful IGC attributions, i.e.\ a \textit{stable} localization of input information for each task under scrutiny.

Finally, through synthetic and practical applications, we demonstrated that IGC attributions reveal selective patterns, coherent with respective model objectives.

%-------------------------------- Back Matter

\section*{Acknowledgements}

Supported by the National Science and Technology Council (NSTC). Collection of the NSD dataset was supported by NSF IIS-1822683 and NSF IIS-1822929.

\section*{Impact Statement}

Our method is primarily designed to improve deep models interpretability, and it does not provide direct generative abilities, nor misuses we can identify. As result, we did not address further societal impact in the paper.

%----------------

\bibliography{biblio}
\bibliographystyle{icml2025}

%----------------

\newpage
\appendix
\onecolumn

\section{Appendix}
\renewcommand{\thefigure}{A\arabic{figure}}
\renewcommand{\theequation}{A\arabic{equation}}
\renewcommand{\thetable}{A\arabic{table}}

%----------------

\subsection{\textit{Linearity} axiom of \textit{path methods}}\label{sec:linearity}

If we define a supra-model as a weighted sum of sub-models, the \textit{linearity} axiom guarantees that the attributions of the supra-model are a weighted sum of sub-models attributions (with identical weights). For $\alpha,\beta\in\R$ and $f,g\in\gF$, we have:
\begin{equation}
    a_j(\alpha f + \beta g,\gamma,\vx^{(i)},\bar{\vx}) = \alpha\ a_j(f,\gamma,\vx^{(i)},\bar{\vx}) + \beta\ a_j(g,\gamma,\vx^{(i)},\bar{\vx})
\end{equation}

%----------------

\subsection{Proof of IGC property: \textit{completeness to a model prediction score}}\label{sec:igc_completeness_proof}

Using the \textit{completeness} axiom of IG (see Eq.\ref{eq:completeness}), we have:
\begin{equation}
\begin{aligned}
    \sum_{j=1}^m \vb_j
        &= \sum_{j=1}^m \frac{1}{\sigma_{f(\rvx)}\sigma_\ry} \E_{(\rvx,\ry)\sim\dataset}\left[\va_j^{(i)}\times(y^{(i)}-\mu_{\ry})\right]\\
        &= \frac{1}{\sigma_{f(\rvx)}\sigma_\ry} \E_{(\rvx,\ry)\sim\dataset}\left[(y^{(i)}-\mu_{\ry})\sum_{j=1}^m\va_j^{(i)}\right]\\
        &= \frac{1}{\sigma_{f(\rvx)}\sigma_\ry} \E_{(\rvx,\ry)\sim\dataset}\left[(y^{(i)}-\mu_{\ry})\big(f(\vx^{(i)})-f(\bar{\vx})\big)\right]\\
        &= \frac{1}{\sigma_{f(\rvx)}\sigma_\ry} \left(\E_{(\rvx,\ry)\sim\dataset}\left[y^{(i)}\times f(\vx^{(i)})\right] - \cancel{f(\bar{\vx})\E_{\ry\sim\dataset}\left[y^{(i)}\right]} - \mu_\ry\E_{\rvx\sim\dataset}\left[f(\vx^{(i)})\right] + \cancel{\mu_\ry f(\bar{\vx})}\right)\\
        &= \frac{\E_{(\rvx,\ry)\sim\dataset}\left[f(\vx^{(i)})\times y^{(i)}\right] - \mu_{f(\rvx)}\mu_\ry}{\sigma_{f(\rvx)}\sigma_\ry}\\
        &= \rho\big(f(\rvx),\ry\big)
\end{aligned}
\end{equation}

%----------------

\subsection{Benchmark images}\label{sec:benchmark_images}

\begin{figure*}[b]
    \centering
    \includegraphics[width=\linewidth]{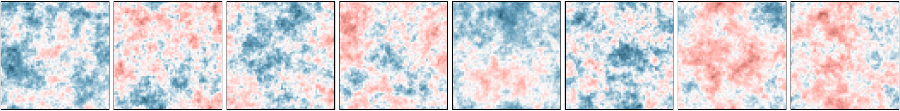}
    \caption{Samples of generated images used in the benchmark experiments presented in Section.\ref{sec:benchmark}. Blue/red pixels indicate negative/positive values.}
    \label{fig:benchmark_images}
\end{figure*}

Fig.\ref{fig:benchmark_images} shows a series of generated images used in the benchmark experiments presented in Section.\ref{sec:benchmark}. These images are designed to respect a known property of natural images, i.e.\ a negative linear correspondence between log-intensities and spatial log-frequencies measured around -1.2 \citep{TolhurstAmplitudespectranatural1992}. This way, while being random, these images exhibit a certain level of spatial redundancy and make the experiments feasible.

The generative procedure is as follows. We first build a 2-d gain map of the target image size filled with the spatial frequencies of the domain used by the selected Fast Fourier Transform (FFT) algorithm. We then take the norm of these cardinal frequencies and raise them to the power of -1.2. Next, resulting gain map is employed to scale complex random maps where each pixel is independently sampled from standard normal distributions for the real and imaginary parts. An inverse FFT finally generates the experimental images.

%----------------

\subsection{Additional benchmark propositions}\label{sec:benchmark_supp}

\begin{figure*}[t]
    \centering
    \includegraphics[width=\linewidth]{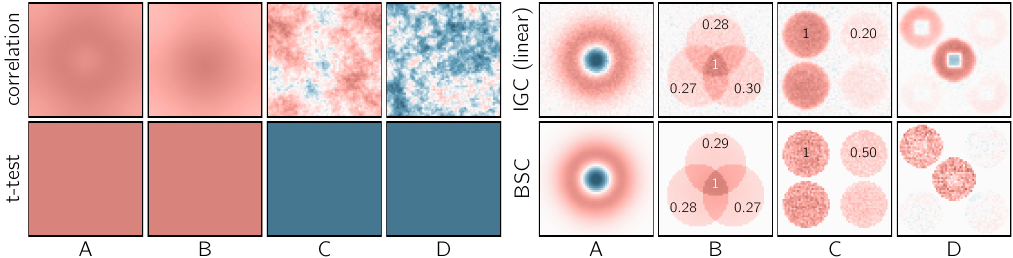}
    \caption{Benchmark of different propositions of \textit{dataset-wise attribution methods} on four experiments labeled A-D. Attribution maps are individually scaled for a better visualization. Blue/red pixels indicate negative/positive values (except for t-tests where red pixels mean $p<0.001$). See Section.\ref{sec:benchmark} for details about the experiments, and the main text for a discussion of displayed propositions.}
    \label{fig:benchmark_supp}
\end{figure*}

In Fig.\ref{fig:benchmark_supp}, we extend the work presented in Section.\ref{sec:benchmark} with other propositions of \textit{dataset-wise attribution methods}. We first report two basic metrics not involving any model. Per-pixel correlation between input components $\rvx_j$ and output $\ry$ provide very blurred versions of expected patterns (Exp.A,B) or complete noise (Exp.C,D). Then, we display two-sample t-tests between input components $\rvx_j$ corresponding to the first and the last decile of $\ry$ statistics for Exp.A,B,C, and categories (one against all others) for Exp.D. Resulting maps are not interpretable because all pixels have the same attribution: all significant (Exp.A,B), i.e. it is possible to discriminate between $\rvx_j$ corresponding to different groups based on $\ry$ ($p<0.001$ in red); all non-significant (Exp.C,D). Together, these two results show that such simple metrics, widely used in neuroscience, do not provide selective and relevant patterns.

Secondly, we explained in Section.\ref{sec:dw_attr_meth} that our method inherits from IG the \textit{implementation invariance} axiom. Here, we investigate the robustness of IGC results upon architectural changes. Unlike \textit{IGC (model)} (see Fig.\ref{fig:benchmark_main}), \textit{IGC (linear)} (see Fig.\ref{fig:benchmark_supp}) does not employ convolutional layers, but only linear layers of sizes (256, 128, 64, 32, 16). Other architectural aspects follow the logic described in Appendix.\ref{sec:benchmark_model_details}. Trained multilinear models still present comparable performance with R2 scores of $>0.99$, $>0.99$, $0.78$ for Exp.A,B,C, and a Top-1 accuracy of 0.76 for Exp.D. The absence of convolutional layers only makes slightly noisier attribution patterns without kernel artifacts.

Finally, we present the results of our method with Baseline Shapley as a supporting \textit{path method} instead of Integrated Gradients. BS is slow to compute, so for practical reasons, we used only 10\% of original samples. Despite a higher noise level, differences with IGC attributions are negligible.

%----------------

\subsection{Details of benchmark models}\label{sec:benchmark_model_details}

\paragraph{IGC computation details}

Our implementation of IGC \citep{LelievreIntegratedGradientCorrelation2025} naturally gives control over the number of IG steps and baselines. These values have been adjusted given the arbitrary rules detailed in Subsection.\ref{sec:practical_igc}. Final parameters are indicated in Table.\ref{tab:benchmark_igc_details}, together with IG/IGC errors and other details. Our implementation is also designed to maximize available resources. To this end, we can adjust the batch size of output components, baselines, and inputs. Presented experiments are relatively simple, so that all output components and baselines have been processed in one batch for 10 inputs. Computation times are reported per output component. Concerning IGC attributions computed on the model, \textit{IGC (model)}, to minimize the noise inherent to the different seeds ruling the weights' initialization and the batch collection, we trained 10 models with different seeds and finally averaged the results of the 5 best ones.

\paragraph{Architectural and training details}

We report the architecture (Table.\ref{tab:benchmark_arch}) and the training details (Table.\ref{tab:benchmark_training_details}) of the models used in the benchmark section. By default, linear layers do not use \textit{bias} parameters because of the use of batch normalization layers (abbreviated b.n.\ in tables). A \textit{bias} is enabled only if it is stated explicitly. Reported tables summarize most of the parameters, but some additional information is required for reproducibility. First, ConvNeXt stem and block units follow the initial definition found in \citet{LiuConvNet2020s2022}. The two exceptions are the stem kernel size limited to 2, and the number of sub-block repetitions per block fixed to 3. The stochastic depth parameter is set to 0.5. Then, the weights of layers followed by Mish activation functions are initialized with normal distributions using Kaiming scaling \citep{HeDelvingDeepRectifiers2015}. Otherwise, we use normal distributions with Xavier scaling \citep{GlorotUnderstandingdifficultytraining2010}. Furthermore, input images and output statistics are all standardized. In the case of images, this is done globally over all pixels. 

%----------------

\subsection{Details of application models}\label{sec:application_model_details}

\paragraph{IGC computation details}

Table.\ref{tab:application_igc_details} provides similar information as in the previous section. The only difference is that computation times (and validation sizes) are reported for the subject 1 of the NSD dataset and for all output components, i.e.\ 2 image statistics and 2973 vertices respectively. The number of output components for the fMRI encoding model is therefore higher by several folds compared to previous applications, and IGC attributions had to be computed for a single input, a single baseline and a limited number of output components per batch. We can notice that corresponding IG error is higher than our arbitrary target of $1\e{-3}$ (see Subsection.\ref{sec:practical_igc}). We could have increased the number of IG steps, but as the IGC error remains at $1\e{-2}$, it did not worth the associated computational cost. For the fMRI decoding model, we also minimized the noise inherent to the different seeds by training 10 different models and averaging the results of the 5 best ones. However, for the fMRI encoding model, we only selected the best one because of the computation time.

\paragraph{Architectural and training details}

We report the architecture (Table.\ref{tab:application_arch}) and the training details (Table.\ref{tab:application_training_details}) of the models used in the application section. Information provided in the previous section still holds for these new models. Missing details concern image and fMRI data processing. Image pixel values are rescaled in range $[0, 1]$ and linearized by inverting the default gamma associated with the sRGB format of png files. We then convert color values to luminance using the Y component of the XYZ-D65 color space and express them on a $log_2$ scale. Finally, images are data augmented at training during the down sampling step. They are first resampled at double size, and then, one pixel is randomly chosen for each 2x2 local cluster. Concerning fMRI data, the standardization is applied per subject and vertex. Next, to achieve a better signal-to-noise ratio of brain activations, the NSD dataset initially provides pre-processed data that are the average of up to three presentations of the same image to the same subject. This is appropriate for the validation set, but not optimal at training. An increased data diversity usually helps models to generalize better. As a result, we used \textit{single} fMRI data at training as a \textit{natural} data augmentation technique. Finally, validation sets correspond to entries of image stimuli shared between subjects.

%---------------- Put all tables at the end

\begin{table}[p]
    \caption{IGC computation details (benchmark)}
    \label{tab:benchmark_igc_details}
    \centering
    \begin{tabular}{lllllllll}
        \toprule
                            & \multicolumn{4}{c}{IGC}                           & \multicolumn{4}{c}{IGC (model)}                   \\
                              \cmidrule(r){2-5}                                   \cmidrule(r){6-9}
                            & Exp.A      & Exp.B      & Exp.C      & Exp.D      & Exp.A      & Exp.B      & Exp.C      & Exp.D      \\
        \midrule
        GPU                 & \multicolumn{8}{c}{NVIDIA RTX A2000 12GB}                                                             \\
        validation size     & 100k       & 100k       & 100k       & 100k       & 10k        & 10k        & 10k        & 10k        \\
        IG n steps          & 512        & 512        & 512        & 512        & 64         & 64         & 64         & 64         \\
        n baselines         & 8          & 8          & 8          & 8          & 8          & 8          & 8          & 8          \\
        IG error            & $<1\e{-6}$ & $2\e{-4}$  & $9\e{-4}$  & $1\e{-5}$  & $1\e{-4}$  & $1\e{-4}$  & $1\e{-4}$  & $1\e{-4}$  \\
        IGC error           & $5\e{-4}$  & $3\e{-4}$  & $3\e{-3}$  & $2\e{-3}$  & $2\e{-3}$  & $2\e{-3}$  & $2\e{-3}$  & $2\e{-3}$  \\
        computation time    & 42.2 m     & 1.4 h      & 1.4 h      & 1.8 h      & 23.9 m     & 24.3 m     & 24.4 m     & 26.3 m     \\
        \bottomrule
    \end{tabular}
\end{table}

\begin{table}[p]
    \caption{Model architectures (benchmark)}
    \label{tab:benchmark_arch}
    \centering
    \begin{tabular}{lll}
        \toprule
                    & Exp.A,B,C           & Exp.D               \\
        \midrule
        input       & images              & images              \\
        input size  & 64$\times$64        & 64$\times$64        \\
        \midrule
        layers      & ConvNeXt stem  16   & ConvNeXt stem  16   \\
                    & ConvNeXt block 32   & ConvNeXt block 32   \\
                    & ConvNeXt block 64   & ConvNeXt block 64   \\
                    & ConvNeXt block 128  & ConvNeXt block 128  \\
                    & ConvNeXt block 256  & ConvNeXt block 256  \\
                    & flatten             & flatten             \\
                    & linear 128          & linear 128          \\
                    & + b.n. + Mish       & + b.n. + Mish       \\
                    & + dropout 0.25      & + dropout 0.25      \\
                    & linear 16           & linear 16           \\
                    & + b.n. + Mish       & + b.n. + Mish       \\
                    & + dropout 0.25      & + dropout 0.25      \\
                    & linear 1 bias       & linear 4 bias       \\
                    &                     & + log softmax       \\
        \midrule
        output      & localized statistic & categories          \\
        output size & 1                   & 4                   \\
        \bottomrule
    \end{tabular}
\end{table}

\begin{table}[p]
    \caption{Training details (benchmark)}
    \label{tab:benchmark_training_details}
    \centering
    \begin{tabular}{lll}
        \toprule
                          & Exp.A,B,C            & Exp.D                \\
        \midrule
        train/val. split  & 100k / 10k           & 100k / 10k           \\
        training epochs   & 50                   & 100                  \\
        batch size        & 64                   & 64                   \\
        n workers (batch) & 16                   & 8                    \\
        optimizer         & \multicolumn{2}{c}{Adam\citep{KingmaAdamMethodStochastic2017}} \\
        learning rate     & $5\e{-5}$            & $5\e{-5}$            \\
        scheduler         & \multicolumn{2}{c}{reduce learning rate on plateau} \\
        sched. decay      & 0.9                  & 0.9                  \\
        sched. patience   & 4                    & 4                    \\
        min learn. rate   & $1\e{-6}$            & $1\e{-6}$            \\
        \midrule
        loss function     & MSE                  & NLL                  \\
        computation time  & 53.1 m               & 2.0 h               \\
        \bottomrule
    \end{tabular}
\end{table}

%----------------

\begin{table}[p]
    \caption{IGC computation details (applications)}
    \label{tab:application_igc_details}
    \centering
    \begin{tabular}{lll}
        \toprule
                            & \multicolumn{2}{c}{models}\\
                              \cmidrule(r){2-3}
                            & \hyperref[sec:fmri_dec]{fMRI decoding} & \hyperref[sec:fmri_enc]{fMRI encoding}\\
        \midrule
        GPU                 & \multicolumn{2}{c}{NVIDIA RTX A2000 12GB}                                      \\
        validation size     & 1000                                   & 1000                                  \\
        IG n steps          & 64                                     & 64                                    \\
        n baselines         & 12                                     & 6                                     \\
        IG error            & $2\e{-5}$                              & $1.5\e{-3}$                           \\
        IGC error           & $8\e{-3}$                              & $1\e{-2}$                             \\
        computation time    & 1.9 m                                  & 73.7 h                                \\
        \bottomrule
    \end{tabular}
\end{table}

\begin{table}[p]
    \caption{Model architectures (applications)}
    \label{tab:application_arch}
    \centering
    \begin{tabular}{lll}
        \toprule
                    & \hyperref[sec:fmri_dec]{fMRI decoding} & \hyperref[sec:fmri_enc]{fMRI encoding}\\
        \midrule
        input       & fMRI data                              & images                                \\
        input size  & 39548                                  & 64$\times$64$\times$3                 \\
        \midrule
        layers      & dropout 0.25                           & ConvNeXt stem  32                     \\
                    & linear 128     ($\times$ 8 subjects)   & ConvNeXt block 64                     \\
                    & + b.n. + Mish  ($\times$ 8 subjects)   & ConvNeXt block 128                    \\
                    & + dropout 0.25 ($\times$ 8 subjects)   & ConvNeXt block 256                    \\
                    & linear 64                              & ConvNeXt block 512                    \\
                    & + b.n. + Mish                          & flatten                               \\
                    & + dropout 0.25                         & linear 256                            \\
                    & linear 32                              & + b.n. + Mish                         \\
                    & + b.n. + Mish                          & + dropout 0.25                        \\
                    & + dropout 0.25                         & linear 4321 bias ($\times$ 8 subjects)\\
                    & linear 16                              &                                       \\
                    & + b.n. + Mish                          &                                       \\
                    & + dropout 0.25                         &                                       \\
                    & linear 8                               &                                       \\
                    & + b.n. + Mish                          &                                       \\
                    & linear 2 bias                          &                                       \\
        \midrule
        output      & image statistics                       & fMRI data                             \\
        output size & 2                                      & 4321                                  \\
        \bottomrule
    \end{tabular}
\end{table}

\begin{table}[p]
    \caption{Training details (applications)}
    \label{tab:application_training_details}
    \centering
    \begin{tabular}{lll}
        \toprule
                          & \hyperref[sec:fmri_dec]{fMRI decoding} & \hyperref[sec:fmri_enc]{fMRI encoding}\\
        \midrule
        train/val. split  & 191882 / 7674                          & 191882 / 7674                         \\
        training epochs   & 75                                     & 50                                    \\
        batch size        & 8 ($\times$ 8 subjects)                & 4 ($\times$ 8 subjects)               \\
        n workers (batch) & 1 ($\times$ 8 subjects)                & 2 ($\times$ 8 subjects)               \\
        optimizer         & \multicolumn{2}{c}{Adam\citep{KingmaAdamMethodStochastic2017}}                 \\
        learning rate     & $1\e{-4}$                              & $1\e{-4}$                             \\
        scheduler         & \multicolumn{2}{c}{reduce learning rate on plateau} \\
        sched. decay      & 0.9                                    & 0.9                                   \\
        sched. patience   & 4                                      & 4                                     \\
        min learn. rate   & $1\e{-6}$                              & $1\e{-6}$                             \\
        \midrule
        loss function     & MSE                                    & MSE                                   \\
        computation time  & 2.4 h                                  & 7.4 h                                 \\
        \bottomrule
    \end{tabular}
\end{table}

\end{document}